\documentclass[10pt,twocolumn,letterpaper]{article}

\usepackage[pagenumbers]{cvpr} 

\usepackage[dvipsnames]{xcolor}

\usepackage[utf8]{inputenc} 
\usepackage[T1]{fontenc}    
\usepackage{booktabs}       
\usepackage{amsfonts}       
\usepackage{nicefrac}       
\usepackage{fancyhdr}       
\usepackage{graphicx}       
\usepackage{amsmath} 
\usepackage{amssymb}  
\usepackage{amsthm}
\usepackage{soul}
\usepackage{balance}
\usepackage{flushend}
\usepackage{array}
\usepackage{multirow}
\usepackage{pifont}
\usepackage{bbold}
\usepackage{nicefrac}
\usepackage{algorithm}
\usepackage{algpseudocode}
\usepackage{wrapfig, calc}

\usepackage[accsupp]{axessibility}  

\DeclareMathOperator*{\argmax}{arg\,max}  
\DeclareMathOperator*{\argmin}{arg\,min}  

\definecolor{cvprblue}{rgb}{0.21,0.49,0.74}
\usepackage[pagebackref,breaklinks,colorlinks,citecolor=cvprblue]{hyperref}

\usepackage[capitalize]{cleveref}

\def\paperID{*****} 
\def\confName{EarthVision}
\def\confYear{2024}

\title{Unsupervised Domain Adaptation Architecture Search with Self-Training for Land Cover Mapping}

\author{Clifford Broni-Bediako$^\ast$\\
\and
Junshi Xia$^\ast$\\
\and
Naoto Yokoya$^{\dagger,\ast}$\\
\and
$^\ast$RIKEN Center for Advanced Intelligence Project (AIP), RIKEN, Tokyo, Japan\\
$^\dagger$Graduate School of Frontier Sciences, The University of Tokyo, Chiba, Japan\\
{\tt\small \{clifford.broni-bediako, junshi.xia\}@riken.jp}, {\tt\small yokoya@k.u-tokyo.ac.jp}
}

\begin{document}
\maketitle

\begin{abstract}
Unsupervised domain adaptation (UDA) is a challenging open problem in land cover mapping. Previous studies show encouraging progress in addressing cross-domain distribution shifts on remote sensing benchmarks for land cover mapping. The existing works are mainly built on large neural network architectures, which makes them resource-hungry systems, limiting their practical impact for many real-world applications in resource-constrained environments. Thus, we proposed a simple yet effective framework to search for lightweight neural networks automatically for land cover mapping tasks under domain shifts. This is achieved by integrating Markov random field neural architecture search (MRF-NAS) into a self-training UDA framework to search for efficient and effective networks under a limited computation budget. This is the first attempt to combine NAS with self-training UDA as a single framework for land cover mapping. We also investigate two different pseudo-labelling approaches (confidence-based and energy-based) in self-training scheme. Experimental results on two recent datasets (OpenEarthMap \& FLAIR \#1) for remote sensing UDA demonstrate a satisfactory performance. With only less than 2M parameters and 30.16 $G$ FLOPs, the best-discovered lightweight network reaches state-of-the-art performance on the regional target domain of OpenEarthMap (59.38\% mIoU) and the considered target domain of FLAIR \#1 (51.19\% mIoU). The code is at \href{https://github.com/cliffbb/UDA-NAS}{https://github.com/cliffbb/UDA-NAS}. 
\end{abstract}

\vspace{-0.2cm}
\section{Introduction}\label{sec:1}
Land cover mapping with very high-resolution optical (VHR) remote sensing (RS) imagery provides detailed information about the Earth's surface for applications such as environmental monitoring \cite{chen2023land, YUAN2020}, urban planning \cite{JIA202497}, disaster response and damage assessment \cite{adriano2021, xia2023damage}, and precision agriculture \cite{SEELAN2003}. With supervised learning methods \cite{richards2022remote}, land cover mapping can be fulfilled automatically by semantic categorization of each pixel in a remotely-sensed optical image. Supervised learning methods assume that the training data and test data have the same distribution, which might not be possible in real-world land cover mapping tasks since the varying geographical regions present vastly different land features with different distributions (referred to as domain-shift) \cite{qui2022}. Thus, well-trained classifiers, for example, deep neural networks, perform poorly on the test data (target domain) that is different from the training data (source domain) distribution \cite{Prasanna2023CVPR}. 

\begin{figure}
    \centering
    \includegraphics[width=1\linewidth]{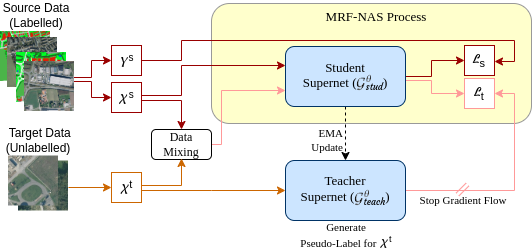}
    \vspace{-0.5cm}
    \caption{Self-training UDA with MRF-NAS process. The \textit{supernet} can be any neural network. After
    the search, i.e., learning pairwise factors in MRF of the student supernet $\mathcal{G}^{\theta}_{stud}$, $m$ optimal subnets (lightweight networks) are inference over the learned factors via diverse M-best loopy inference \cite{payman2012}. Then, 
    we retrained the discovered lightweight networks. $\mathcal{L}_s$ and $\mathcal{L}_t$ are cross-entropy loss functions for the source labelled data and target pseudo-labelled data, respectively.  The teacher supernet $\mathcal{G}^{\theta}_{teach}$ is only used to generate pseudo-labels for the target images to train  $\mathcal{G}^{\theta}_{stud}$, in addition to the source labelled data. See Section \ref{sec:3} for more details. 
    }
    \label{fig:uda-nas-framework}
    \vspace{-0.45cm}
\end{figure}

Annotating a large dataset of a new geographical region (new domain) for supervised learning is costly and time-consuming \cite{garioud2022flair1,junshi2023}. Researchers in the RS community have explored unsupervised domain adaptation (UDA) techniques \cite{mengqui2022UDAsurvey} to adapt neural networks trained on labelled data in one domain (source domain) to a new domain (target domain), which only has unlabelled data. While successful progress has been made in the traditional UDA for land cover mapping \cite{SOTOVEGA2021113,liang2023,yikun2022}, lightweight UDA remains a challenge and under-investigated. The existing UDA methods for land cover mapping rely on handcrafted large neural network architectures  \cite{capliez2023,kapil2024shadowsense,caiyu2022}, which may not be practically suitable for real-world applications across a wide range of resource-constrained platforms such as nanosatellites (CubeSat) \cite{denby2019}, drones \cite{MOLLICK2023}, and mobile devices \cite{melis2018}. 

Marsocci~\textit{et al.,} \cite{marcocci2023} recently proposed GeoMultiTaskNet with 33$M$ parameters to address the lightweight network challenge in UDA land cover mapping tasks. The size of GeoMultiTaskNet might not fit a wide range of environments where computation resources are limited. Thus, it is important to develop lightweight networks for UDA land cover mapping tasks. This will provide an opportunity for a practical land cover mapping domain adaptation deployment in resource-constrained environments. Doing this manually is a daunting exercise, especially, since the optimal networks can be data-dependent across different geographical regions and scenes. 
Thus, we turn to automated neural architecture search (NAS) \cite{thomas2019NAS} to automatically search for lightweight networks with different computation budgets for resource-constrained platforms in UDA land cover mapping tasks. In computer vision, there are some studies on NAS for UDA, but most of the studies have been performed in UDA for image classification \cite{meng2022slimmable, Robbiano2022,sheng2021evolving, Yanxi2020, Yue2022,chakraborty2023anyda}. The challenging UDA for semantic segmentation \cite{deng2021autoadapt} is barely investigated. On the other hand, NAS for UDA in RS remains unexplored. 

To this end, in this paper, we explore lightweight NAS for UDA land cover mapping on two new RS benchmarks, OpenEarthMap~\cite{junshi2023} and FLAIR \#1~\cite{garioud2022flair1} datastes, which have been properly curated for UDA tasks in land cover mapping. Specifically, we extend Markov random fields neural architecture search (MRF-NAS) \cite{Wang2022MRF-UNets} to a UDA scheme to learn a single neural network, called \textit{supernet}, using both source domain labelled data and target domain unlabelled data, and inference over the learnt supernet to find subnets (lightweight networks) for resource-constrained platforms. 
Adapting NAS into UDA is challenging as the target domain has only unlabelled data \cite{Robbiano2022}. Just searching and building networks, and then applying the networks for domain adaptation might not be effective \cite{deng2021autoadapt,chakraborty2023anyda}. It requires different training configurations for the weights and architecture adaptation \cite{meng2022slimmable}. 
Therefore, we propose \textit{self-training UDA-NAS framework} of which the workflow process is shown in Figure~\ref{fig:uda-nas-framework}. 
Self-training UDA employs a teacher network to produce pseudo-labels for a target domain. 
Applying pseudo-labelling naively in UDA can lead to training difficulty for a network to learn the hard-to-predict classes \cite{zou2018unsupervised}. For this reason, in addition to confidence-based pseudo-labelling, which is commonly used in self-training UDA \cite{zhang2021prototypical,tranheden2021dacs,hoyer2022daformer}, we also investigate energy-based pseudo-labelling technique \cite{yu2023inpl} for self-training in the context of UDA-NAS. 

The key contributions of this paper are summarised as follows. We propose a self-training UDA-NAS framework, which extends MRF-NAS to a UDA scheme, to search for lightweight networks for land cover mapping that has been beyond the reach of previous UDA land cover mapping methods. This offers an opportunity to leverage the advancement made in UDA land cover mapping to address the practical problem of deploying UDA in a resource-constrained environment. Moreover, with very few parameters ($<2M$), the discovered lightweight networks improve the performance of the handcrafted large networks on the OpenEarthMap and the FLAIR \#1 datasets. Furthermore, the extensive experiments on the two benchmarks show promising performance of applying confidence-based pseudo-labelling techniques over energy-based pseudo-labelling in lightweight UDA-NAS for land cover mapping. 

\section{Related Work}\label{sec:2}
\subsection{UDA for Land Cover Mapping}\label{sec:2.1}
UDA is employed in land cover mapping to alleviate the need for per-pixel annotation, aiming for the model trained with labelled source data and unlabelled target data to attain high performance on the target data by minimising the dissimilarity between the source and target domains. Many methods, including adversarial learning and self-training, are proposed. Ji \textit{et al.}~\cite{9198144} proposed a generative adversarial network (GAN) based domain adaptation for land cover classification. Furthermore, a category-space-constrained adversarial method was proposed to execute category-level adaptive coastal land cover mapping~\cite{rs13081493}. Colour mapping generative adversarial network (ColorMapGANs) is proposed to reduce the colour gap between source and target domains~\cite{9047180}. Capliez \textit{et al.}~\cite{capliez2023,10089508} adopted adversarial learning, spatially aware self-training, and spatial pseudo-labelling for multi-sensor land cover mapping. Ma~\textit{et al.}~\cite{10094018} proposed novel local consistency and global diversity metrics into the UDA framework. Wang~\textit{et al.}~\cite{WANG2022113058} proposed the LoveCS framework that integrated cross-sensor normalization, self-training domain adaptation, and multi-scale pseudo-labelling. A novel two-stage Domain Adaptation method for Cross-Spatio-Temporal (DASCT) classification was proposed in~\cite{LUO2022105}.  

\subsection{Neural Architecture Search for UDA}\label{sec:2.2}
NAS for UDA leverages automated architecture search to find an optimal network architecture for UDA tasks. There are few studies on NAS for UDA tasks in computer vision, but it remains unexplored in RS image understanding. AdaptNAS~\cite{Yanxi2020} and NASDA~\cite{li2020network} investigated the generalization abilities of NAS for UDA with a gradient-based NAS method. AdaXpert~\cite{niu2021adaxpert} proposed a progressive search to adjust architectures on growing data. Leveraging reinforcement learning, DAMPC-NAS~\cite{yue2021domain} and AASPC~\cite{Yue2022} learned architectures that can align the source and target domains via population correlation. ABAS~\cite{Robbiano2022} learned an auxiliary network branch with Bayesian optimisation in the scheme of adversarial learning UDA. EvoADA~\cite{sheng2021evolving} adopted an evolutionary algorithm to learn domain-conditioned attention mechanisms for UDA tasks. These NAS for UDA methods produce resource-hungry networks for UDA image classification. AutoAdapt\cite{deng2021autoadapt} departed from image classification to search for UDA networks for image semantic segmentation using an evolutionary algorithm with maximum mean discrepancy. However, AutoAdapt also produces resource-hungry UDA networks, which are not suitable for resource-constrained environments. Leveraging knowledge distillation with one-shot NAS technique, SlimDA~\cite{meng2022slimmable} and AnyDA~\cite{chakraborty2023anyda} learned lightweight UDA networks for image classification, but the challenging problem of lightweight UDA for semantic segmentation remains uninvestigated. This paper attempts to address this problem by integrating MRF-NAS into self-training UDA as a single framework to search for lightweight UDA for land cover mapping.

\section{Methodology}\label{sec:3}
The proposed self-training UDA-NAS aims to search for lightweight networks with MRF-NAS strategy by transferring labelled source domain knowledge to an unlabelled target domain. We provide background material in Section~\ref{sec:3.1}. Self-training UDA is described in Section~\ref{sec:3.2} and MRF-NAS is described in Section~\ref{sec:3.3}. Finally, the self-training UDA-NAS framework is illustrated in Section~\ref{sec:3.4}.

\subsection{Preliminaries}\label{sec:3.1}
\paragraph{Neural architecture search (NAS):}
Given a task, NAS aims to search for an optimal network architecture $\alpha^{*}$ automatically, together with its parameters $\theta$, via search strategies such as reinforcement learning \cite{zoph2017neural} and evolutionary algorithms \cite{liuNAS2023}. Without loss of generality, the joint search for optimal architecture and its parameters for a specific task can be formulated as a bi-level optimization problem 
\begin{equation}
\begin{split}
    \alpha^{*} = \argmin_{\alpha\in \mathcal{A}}\mathcal{L}_{search}(\theta^{*}(\alpha), \alpha), \\
    s.t., \:\theta^{*}(\alpha) = \argmin_{\theta}\mathcal{L}_{train}(\theta, \alpha),
\end{split}
\end{equation}
where $\mathcal{A}$ is the search space containing all possible architectures that can be found. 
$\mathcal{L}_{search}$ and $\mathcal{L}_{train}$ indicate validation loss and training loss, respectively, that can be minimised by using a validation set to update $\alpha$ and a training set to update $\theta$. In the one-shot NAS approach \cite{liu2018darts,wu2019fbnet,xie2018snas}, the architecture search space $\mathcal{A}$ is represented as a directed acyclic graph (DAG) and the optimal network architecture is a subgraph $\alpha\in\mathcal{A}$. We assume this definition in this work. Traditionally, NAS methods assume that both the training and the validation sets have labels and are from the same distribution, which is not the case in UDA \cite{liu2022deepUDA,sheng2021evolving,hoyer2022daformer}.

\vspace{-0.3cm}
\paragraph{Unsupervised domain adaptation (UDA):}
Let $\mathcal{S}$ and $\mathcal{T}$ respectively denote a source domain and a target domain that are related semantically. We are given a labelled dataset  $\mathcal{D}_s=\{(\mathbf{x}^{s}_{i}, \mathbf{y}^{s}_{i})\}_{i=1}^{n_s}$ of $n_s$ examples from $\mathcal{S}$ and unlabelled dataset $\mathcal{D}_t=\{\mathbf{x}^{t}_{i}\}_{i=1}^{n_t}$ of $n_t$ examples from $\mathcal{T}$. UDA aims at mitigating the domain gap between $\mathcal{S}$ and $\mathcal{T}$ by training a neural network $f_{\theta}$ on the labelled dataset $\mathcal{D}_s$ to exploit meaningful knowledge learned from $\mathcal{S}$, to perform well on the unlabelled dataset $\mathcal{D}_t$ of $\mathcal{T}$ \cite{liu2022deepUDA,sheng2021evolving,hoyer2022daformer}.

\subsection{Architecture search strategy}\label{sec:3.2}
In this work, we leverage MRF-NAS~\cite{Wang2022MRF-UNets} as the architecture search strategy to build the proposed self-training UDA-NAS framework. The goal is to search for lightweight networks that are optimal to achieve domain alignment between $\mathcal{D}_s$ and $\mathcal{D}_t$. MRF-NAS is a resource-aware architecture search method based on AOWS~\cite{berman2020aows}. It can find architectures that satisfy resource constraints such as the number of floating-point operations (FLOPs) and latency. We briefly describe MRF-NAS and we refer to Wang~\textit{et al.,} \cite{Wang2022MRF-UNets} for more details.

Denote $\mathcal{A}$ as a neural network with $n$ choices of nodes such that $\alpha_{n}=\{n_{i}|i \in [n]\}\}$, node $n_{i}$ takes value from a finite label set $Q_i$ (e.g., kernel size = 3, 5, or 7), and $\mathcal{A(\alpha)}$ as sub-network (subnet) of $\mathcal{A}$ with $\alpha$ architecture. $[n]$ is a short form for $\{1, 2, ..., n\}$.
The MRF-NAS represents the NAS problem as maximum a posteriori (MAP) inference over pairwise MRF \cite{koller2009probabilistic}. 
Let $\Psi$ denotes a set of factors $\{\psi_{S}|S\subseteq[n]\: \text{and}\: |S|\leq 2\}$ in a pairwise MRF of $\mathcal{A}$, 
where $\psi_{S}:\alpha_{S}\rightarrow \mathbb{R}$. 
Let $P_\Psi(\alpha) = \frac{1}{Z}\exp(\sum_{S}\psi_{S}(\alpha_{s}))$ denotes a probabilistic distribution of the set of factors, where $Z$ is the normalizing constant. 
Given a task-specific performance measurement $\mathcal{M}$ (e.g.
classification accuracy),  $P_\Psi(\alpha_1) \geq P_\Psi(\alpha_2) \Rightarrow \mathcal{M}(\mathcal{A}(\alpha_1)) \geq \mathcal{M}(\mathcal{A}(\alpha_2))$. The NAS problem then becomes MAP inference over a set of defined factors, which aims to find the optimal $\alpha^\ast$ as 
\begin{equation}
\begin{split}
    \alpha^\ast = \argmax_\alpha \mathcal{M}(\mathcal{A}(\alpha)) = \argmax_{\alpha}E{(P_\Psi(\alpha))}, 
\end{split}
\end{equation}
where $E$ is an energy function. The factors are learned via differentiable parameter learning \cite{ardywibowo2020,liu2018darts} by minimising the following objective
\begin{equation}\label{equ:3}
    -\mathbb{E}_{\alpha\sim P_{\Psi}}(\alpha)\bigl[\mathcal{L}(\theta^\ast|\alpha)\bigr] \:\:\:s.t.\: \:\theta^\ast=\argmin_{\theta}\mathcal{L}(\theta|\alpha)
\end{equation}
where $\mathcal{L}$ is a cross-entropy loss function.
Monte Carlo approximation (e.g., Gibbs sampling) is applied to make the Equation (\ref{equ:3}) differentiable. Then it is solved by adopting the Gumbel-Softmax reparameterization trick to smooth the discrete categorical distribution \cite{jang2017categorical}.

After learning the factors, the diverse M-best loopy inference method \cite{payman2012} is employed to find a set of diverse optimal subnets (lightweight networks) $\{\alpha_{1}, \alpha_{2}, ..., \alpha_{m}\}$. This makes the MRF-NAS method akin to one-shot NAS methods with weight sharing \cite{bender2018understanding,chu2021fairnas}. The advantage of MRF-NAS is that it imposes factorisation, therefore, the cardinality of its set of factors does not grow exponentially in the order of the number of nodes in the network, as it does in other one-shot NAS methods \cite{chu2021fairnas,bender2018understanding,guo2020single}. 

\subsection{Self-training UDA}\label{sec:3.3}
Most existing UDA methods can be grouped into self-training \cite{tranheden2021dacs,hoyer2022daformer,zou2018unsupervised} and adversarial training \cite{Tsai_adaptseg_2018,luo2019Taking, Haoran_2020_ECCV} approaches. In this work, we employ self-training UDA as adversarial training UDA is less stable, and the self-training methods usually perform better than the adversarial training ones \cite{tranheden2021dacs, hoyer2022daformer}. 
To address the domain gap between $\mathcal{S}$ and $\mathcal{T}$, a self-training UDA method adopts a teacher network $h_{\phi}$ to generate pseudo-labels $\{\hat{\mathbf{y}}^{t}_{i}\}_{i=1}^{n_t}$ for the unlabelled images $\{\mathbf{x}^{t}_{i}\}_{i=1}^{n_t}$ in $\mathcal{D}_t$ as 
\begin{equation}
    p^{t}_{(i,j,c)} =[c= \argmax_{c'} h_{\phi}(\mathbf{x}^{t}_{i})_{(j,c')}],
\end{equation}
where $[\cdot]$ denotes Iverson bracket and $c$ is the class index for pixel $j$ in $\mathbf{x}^{t}_{i}$.  
As adopted in DAFormer~\cite{hoyer2022daformer}, $h_{\phi}$ applies a confidence estimate with threshold $\tau$ of maximum softmax probability \cite{tranheden2021dacs} to generate the pseudo-labels. 
Hence,
\begin{equation}
    \hat{\mathbf{y}}^{t}_{i} = p^{t}_{(i,j,c)}\frac{\sum_{j=1}^{H\times W}[\text{max}_{c'}h_{\phi}(\mathbf{x}^{t}_{i})_{(j,c')} \geq \tau]}{H \times W},
\end{equation}
where $H$ and $W$ are the height and width of $\mathbf{x}^{t}_{i}$, respectively. 

As an alternative, energy score threshold $\tau_{e}$ can also be applied to generate the pseudo-labels \cite{yu2023inpl}. The energy score is defined as \cite{lecun2006tutorial}:
\begin{equation}
    E(\mathbf{x}^{t}_{i}, h_{\phi}(\mathbf{x}^{t}_{i})) = -T\log\Bigl(\sum_{c=1}^{C}e^{\nicefrac{{h_{\phi}(\mathbf{x}^{t}_{i})_{c}}}{T}}\Bigr),
\end{equation}
where ${h_{\phi}(\cdot)_c}$ indicates the corresponding logit value of the $c$-th class, $C$ is the total number of classes, and $T$ is a temperature that can be tuned. 
Applying the energy score, the pseudo-labels are generated as:
\begin{equation}
    \hat{\mathbf{y}}^{t}_{i} = p^{t}_{(i,j,c)}[E(\mathbf{x}^{t}_{i}, h_{\phi}(\mathbf{x}^{t}_{i}))<\tau_{e}].
\end{equation}

We investigate both the confidence-based pseudo-labelling and the energy-based pseudo-labelling techniques. The parameters $\theta$ of a student network $f_{\theta}$ is trained by minimising the following loss: 
\begin{equation}
    \mathcal{L}({\theta}) = \mathbb{E}\bigl[H\bigl(f_{\theta}(\mathcal{X}^{s}),\mathcal{Y}^{s}\bigr) \:+\: \lambda H\bigl(f_{\theta}(\mathcal{X}^{t}),\hat{\mathcal{Y}}^{t}\bigr)\bigr],
\end{equation}
where the expectation $\mathbb{E}$ is over batches of random variables of image-label pairs ($\mathcal{X}^{s},\mathcal{Y}^{s}$) and ($\mathcal{X}^{t},\hat{\mathcal{Y}}^{t}$) sampled uniformly from $\mathcal{D}_s$ and $\mathcal{D}_t$ (with generated pseudo-labels), respectively. $H$ indicates the cross-entropy between the predictions and the reference labels averaged over all images and pixels, and $\lambda$ is a hyperparameter to determine the amount of the loss in $\mathcal{T}$ that affects the overall training. 
Note that the parameters $\phi$ of the teacher network $h_{\phi}$ are not updated via backpropagated gradients. They are updated based on the exponentially moving average of the parameters $\theta$ of the student network $f_{\theta}$ after each training epoch to improve the prediction stability of the teacher network $h_{\phi}$ \cite{tarvainen2017mean}. And we adopt online self-training \cite{hoyer2022daformer,zhou2022context, Araslanov2021DASAC} in generating the pseudo-labels as the alternative offline self-training has a complex training setup \cite{zou2018unsupervised,zou2019confidence}. 

\subsection{UDA-NAS framework}\label{sec:3.4}
Figure~\ref{fig:uda-nas-framework} provides an overview of the proposed self-training UDA-NAS framework. The framework integrates NAS into UDA by applying the MRF-NAS strategy (see Section~\ref{sec:3.2}) for resource-aware search in the self-training UDA scheme (see Section~\ref{sec:3.3}) to find optimal architectures that can satisfy resource-constrained platforms. Based on the teacher-student scheme, our self-training UDA-NAS framework consists of a teacher supernet $\mathcal{G}^{\theta}_{teach}$ (which correspond to teacher network $h_{\phi}$ in Section~\ref{sec:3.3}) and a student supernet $\mathcal{G}^{\theta}_{stud}$ (which correspond to student network $f_{\theta}$ in Section~\ref{sec:3.3}) with the architectures parameterised by $\theta_{teach}$ and $\theta_{stud}$, respectively, and the student supernet $\mathcal{G}^{\theta}_{stud}$ is configured with MRF-NAS scheme. The teacher supernet $\mathcal{G}^{\theta}_{teach}$ is only used to generate pseudo-labels for the unlabelled images in $\mathcal{D}_t$, while the architecture search takes place in the student supernet $\mathcal{G}^{\theta}_{stud}$ space only. 

The student supernet $\mathcal{G}^{\theta}_{stud}$
takes both the source and target images to learn architectures, whereas the teacher supernet $\mathcal{G}^{\theta}_{teach}$ is fed with only the target images to generate their pseudo-labels. During the search, that is, learning pairwise and unary factors in the MRF of $\mathcal{G}^{\theta}_{stud}$ with differentiable parameter learning \cite{ardywibowo2020,liu2018darts}, $\mathcal{G}^{\theta}_{stud}$ is trained based on supervised learning using both $\mathcal{D}_s$ and $\mathcal{D}_t$ (with the pseudo-labels). $\mathcal{G}^{\theta}_{stud}$ is trained with knowledge obtained from $\mathcal{G}^{\theta}_{teach}$, to enable its subnets (lightweight networks) capable of aligning the source and the target domains under limited computation budget. Following DACS~\cite{tranheden2021dacs}, the ClassMix \cite{olsson2021classmix} mixing strategy is applied to mix the images from $\mathcal{D}_s$ and $\mathcal{D}_t$ to enable the subnets of $\mathcal{G}^{\theta}_{stud}$ to learn cross-domain robust features. The parameters $\theta_{teach}$ of $\mathcal{G}^{\theta}_{teach}$ are updated via the exponential moving average of the parameters $\theta_{stud}$ of  $\mathcal{G}^{\theta}_{stud}$. 

Then, after the search, the diverse M-best loopy inference \cite{payman2012} is applied over the learned $\mathcal{G}^{\theta}_{stud}$ MRF factors to find $m$ optimal subnets (lightweight networks) that achieve good performance under a given computation budget. Finally, the discovered lightweight networks are retrained, as normally done in NAS \cite{deng2021autoadapt,bbc2022,zoph2018learning}, in the self-training UDA mode (without the MFR-NAS process) to find the best optimal network.

\section{Experimental Setup}\label{sec:4}
\subsection{Datasets}\label{sec:4.1}
\textbf{OpenEarthMap:} OpenEarthMap serves as a benchmark dataset dedicated to advancing global high-resolution land cover mapping. It encompasses 5000 aerial and satellite images with the size of $1024 \times 1024$, each manually annotated with 8-class land cover labels. Covering 97 regions across 44 countries on 6 continents, OpenEarthMap comprises 2.2 million segments captured at ground sampling distances ranging from 0.25 to 0.5 meters~\cite{junshi2023}. For the UDA setting, the dataset is divided into 73 regions for the source domain and 24 regions for the target domain. This division ensures a balanced representation across both domains, encompassing countries from all six continents and maintaining an equilibrium between urban and rural areas.\\
\textbf{FLAIR \#1:} The French Land cover from Aerospace ImageRy dataset (FLAIR) comprises 50 spatial domains, each representing the varying landscapes and climates of metropolitan France, with each domain corresponding to a French department~\cite{garioud2022flair1}. 
Each patch measures $512 \times 512$ pixels, with a GSD of 0.2 $m$, and each domain consists of 1725 to 1800 patches. Following the experimental settings in GeoMultiTaskNet~\cite{marcocci2023}, ten (D06, D08, D13, D17, D23, D29, D33, D58, D67, D74) and three (D64, D68, D71) departments are selected as the source and target domains, respectively, consisting of 16k images for training and 5k
for testing with 12 classes. 

\subsection{Baselines}\label{sec:4.2}
Unfortunately, we could not compare the performance of the proposed method to any UDA-NAS methods for land cover mapping as the code for 
the previous UDA-NAS method for semantic segmentation \cite{deng2021autoadapt} is unavailable. However, we compare our work to 9 handcrafted, state-of-the-art, semantic segmentation UDA methods for land cover mapping. These are: AdaptSegNet~\cite{Tsai_adaptseg_2018}, CLAN~\cite{luo2019Taking}, TransNorm~\cite{Wang19TransNorm}, CBST~\cite{zou2018unsupervised}, and IAST~\cite{mei2020instance} on the OpenEarthMap dataset (see Table~\ref{tab:oem_search_results}) and AdaptSegNet~\cite{Tsai_adaptseg_2018}, ADVENT~\cite{vu2018advent}, DAFormer~\cite{hoyer2022daformer}, UDA\_for\_RS~\cite{li2022}, and GeoMultiTaskNet~\cite{marcocci2023} on the FLAIR \#1 dataset (see Table~\ref{tab:flair_transfer_results}).

\subsection{Implementation details}\label{sec:4.3}
All the experiments are implemented in the PyTorch library and run on a single NVIDIA Tesla P100 (DGX-2) with 32GB memory. The search is performed on the OpenEarthMap dataset.  We evaluate the discovered architectures on the FLAIR \#1 dataset as well. 

\vspace{-0.3cm}
\paragraph{Search space:}
The search space contains all the possible architectures that can be found. 
In this work, we adopt the same search space provided in Wang~\textit{et al.} \cite{Wang2022MRF-UNets} that was designed on the U-Net architecture \cite{ronneberger2015u} with $4\times 10^{23}$ configurations. The search space has three neural network operations with different kernel sizes and different width ratios of the network as presented in Table~\ref{tab:search-space}. 

\begin{table}[!h]
\centering
\scalebox{1}{
\begin{tabular}{l c c}
    \toprule
    {Operation}	& {Size} & {Width} \\ 
    \toprule
    Normal & 3, 5 & 0.5, 0.75, 1.0, 1.25, 1.5 \\
    Downsampling & 3 & 0.5, 0.75, 1.0, 1.25, 1.5 \\
    Upsampling & 2 & 0.5, 0.75, 1.0, 1.25, 1.5 \\
    \bottomrule
\end{tabular}}
\captionsetup{width=0.98\linewidth}
\caption{The search space based on U-Net architecture.}
\label{tab:search-space}
\vspace{-0.8cm}
\end{table}

\paragraph{Searching and training:}
During the architecture search, the student supernet $\mathcal{G}^{\theta}_{stud}$ is trained for 40,000 iterations with an initial warmup period of 1,500 iterations before the factors are updated. We employ the sandwich training scheme \cite{yu2019universally} to train $\mathcal{G}^{\theta}_{stud}$ at the largest width, smallest width, and other randomly sampled widths altogether in each iteration. We use the AdamW optimizer with a learning rate of 0.003 for the network weights and a weight decay of 0.05. To improve class imbalance, we adopt recall cross-entropy loss \cite{tian2022}. Following DAFormer~\cite{hoyer2022daformer}, colour jitter and Gaussian blur are applied as data augmentation after a random crop of size $512\times 512$ is applied. Other search settings such as the learning rate for the MRF factors and the hyperparameters for sampling and diverse M-best inference remain the same as in Wang~\textit{et al.} \cite{Wang2022MRF-UNets}. The architecture search is performed in two different pseudo-labelling scenarios of the teacher supernet $\mathcal{G}^{\theta}_{teach}$: (1) applying confidence-based pseudo-labelling with confidence threshold $\tau=0.968$ as in DAFormer~\cite{hoyer2022daformer} and (2) applying energy-based pseudo-labelling with cutoff threshold $\tau_e=-8.0$ and temperature $T=1$ as in InPL~\cite{yu2023inpl}, to generate pseudo-labels for training $\mathcal{G}^{\theta}_{stud}$. For the inference, we select 8 optimal subnet architectures, 4 from each pseudo-labelling scheme, based on a computation budget of 2.5$G$ FLOPs computed on $256\times 256$ image resolution (smaller resolution is used for fast inference). We retrain them and select the top 2 networks from each set as the best-discovered networks. We use the same hyperparameters as in the search phase for the retraining phase, except that we train for 140,000 iterations.

\begin{table*}[t]
\addtolength{\tabcolsep}{0pt}
\centering
\scalebox{0.76}{
\begin{tabular}{l c c c m{1.2cm} m{1cm} m{1cm} c c c c c c}
        \toprule
        \multirow{2}{*}{Method} & \multicolumn{8}{c}{IoU (\%)}  & {mIoU} & {Params} & {FLOPs} & {Speed} \\ 
        \cline{2-9}
        &  Bareland    &	Rangeland	&	Developed	&	Road	&	Tree	&	Water	&	Agriculture	&	Building & (\%) $\uparrow$ & (M) $\downarrow$ & (G) $\downarrow$ & (FPS) $\uparrow$ \\	
        \toprule
        AdaptSegNet~\cite{Tsai_adaptseg_2018} & 28.77&	41.47&	36.09&	45.16&	46.65&	34.48&	68.47&	63.74&	45.60 &30.05 & 134.80 & 9.17\\
        CLAN~\cite{luo2019Taking} & 22.90& 42.25&	39.49&	44.12&	58.98&	58.99&	59.51&	64.53&	48.85 &27.61 & 136.16 & 19.06\\
        TransNorm~\cite{Wang19TransNorm} & 27.54&	45.13&	37.99&	45.56&	57.06&	63.84&	66.26&	64.71&	51.01 &30.38 & 135.68 & 2.23\\
        CBST~\cite{zou2018unsupervised}	& \underline{29.64}& 43.79&	37.99&	49.19&	57.33&	60.75& \underline{71.93}&	65.46&	52.01 &46.59 & 208.18 & \underline{40.28} \\
        IAST~\cite{mei2020instance}	& \bf{33.68}& 43.64&	37.03&	45.16&	59.61&	\underline{72.08} &	\bf{74.72}&	61.77&	53.46 &49.36 & 209.24 & \bf{41.65}\\ 
        \midrule
        Our Net-C1 & 28.00 & \bf{49.41} & \bf{50.35} & 58.70 & \bf{68.75} & \bf{73.51} & 69.20 & \bf{77.20} & \bf{59.38} & \bf{1.28} & \underline{30.16} & 21.79 \\ 
        Our Net-C2 & 26.41 & 49.11 & \underline{50.22} & \bf{58.99} & 67.94 & 70.38 & 69.94 & \underline{77.10} & \underline{58.76} & 1.63 & 30.53 & 22.53 \\
        Our Net-E1 & 22.99 & \underline{49.14} & 48.94 & \underline{58.74} & 67.07 & 69.28 & 67.10 & 74.01 & 57.15 & 1.65 & 33.59 & 21.47\\ 
        Our Net-E2 & 24.25 & 47.60 & 47.91 & 57.45 & \underline{68.21} & 66.17 & 68.90 & 76.40 & 57.11 & \underline{1.45} & \bf{29.16} & 22.39 \\
        \bottomrule
\end{tabular}}
\vspace{-1mm}
\captionsetup{width=0.97\linewidth}
\caption{Comparison with other UDA methods on the regional target domain test set of the OpenEarthMap dataset. In addition to the improved performance in terms of mIoU, the discovered lightweight networks are significantly smaller than that of the handcrafted baselines. The results of the baselines are based on the experiments we performed using their official implementations from GitHub. The best score is in \textbf{bold} and the second-best is \underline{underlined}. }
\label{tab:oem_search_results}
\vspace{0.1cm}
\end{table*}

\begin{figure*}[!t]
\centering
    \includegraphics[width=1\linewidth]{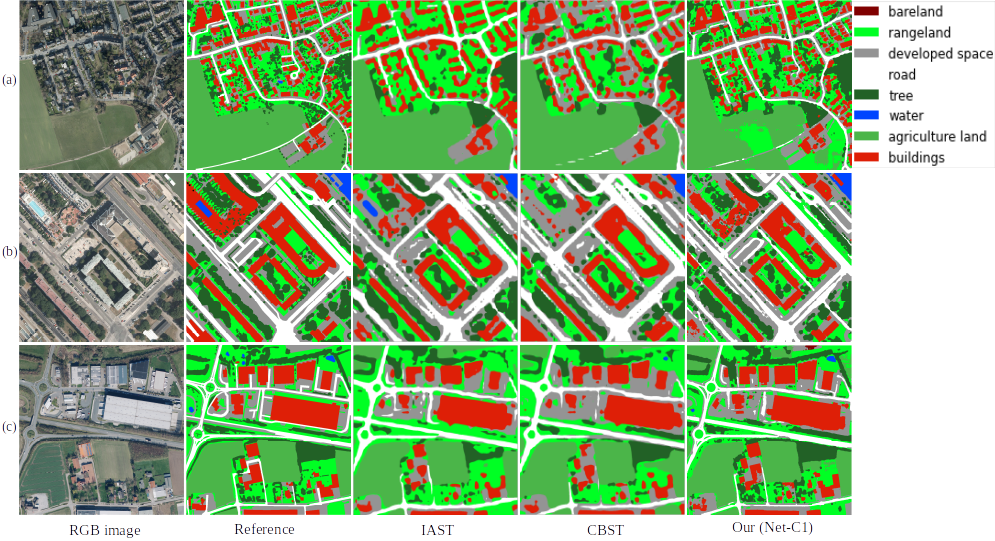}
    \vspace{-5mm}
    \captionsetup{width=0.98\linewidth}
\caption{Visual comparison of land cover mapping results of the best-discovered network and some representative baselines in Table~\ref{tab:oem_search_results}.}
\label{fig:oem_uda_visual}
\vspace{-0.1cm}
\end{figure*}

\section{Experimental Results}\label{sec:5}
Here, we present and discuss the experimental results of the proposed method. Table~\ref{tab:oem_search_results} presents the evaluation of the discovered lightweight networks against UDA baseline methods in terms of accuracy (mIoU) on the OpenEarthMap regional target domain, the number of parameters (M), the FLOPs (G), and the latency (speed in frames per second (FPS)). The results of the baselines shown in Table~\ref{tab:oem_search_results} are based on the experiments we performed using the official implementations obtained from their GitHub sites. In addition, to evaluate the transferability of the discovered networks, we employ the best-discovered networks on the considered target domain of the FLAIR \#1 dataset and compared the results with other UDA methods that were reported in GeoMultiTaskNet~\cite{marcocci2023} (see Table~\ref{tab:flair_transfer_results}). The networks discovered with confidence-based pseudo-labelling are denoted as Net-C and the ones with energy-based pseudo-labelling are denoted as Net-E.

\begin{table*}
\centering
\scalebox{0.72}{
\begin{tabular}{l m{1cm} m{1cm} m{1cm} m{1cm} m{1cm} m{1cm} m{1cm} m{1cm} m{1cm} m{1cm} m{1cm} m{1cm} c c }
    \toprule
    \multirow{2}{*}{Method}	& \multicolumn{12}{c}{IoU (\%)} & {mIoU} & {Params} \\ 
    \cline{2-13}
    & Build & Perv. & Imp. & Bare & Water & Conif. & Decid. & Brush & Vine & Herb & Agric & Plowed & (\%) $\uparrow$ & (M) $\downarrow$ \\
    \toprule
    AdaptSegNet~\cite{Tsai_adaptseg_2018} & 39.98 & 20.75 & 40.23 & 20.36 & 15.25 & 4.93 & 35.37 & 10.99 & 34.51 & 42.69 & 11.06 & 23.47 & 24.97 & 99 \\
    ADVENT~\cite{vu2018advent} & 35.79 & 24.38 & 48.82 & 6.85 & 31.98 & 0.00 & 51.65 & 11.79 & 33.33 & 25.76 & 11.46 & 24.29 & 25.56 & 99 \\
    DAFormer~\cite{hoyer2022daformer} & 67.09 & 45.56 & 61.99 & \underline{55.35} & 65.12 & 8.91 & 54.39 & 20.31 & \underline{64.39} & 38.79 & 23.74 & 41.83 & 45.61 & 85 \\
    UDA\_for\_RS~\cite{li2022} & 66.30 & 48.05 & 62.36 & \bf{59.28} & 61.24 & 9.22 & 60.02 & 16.52 & 57.74 & 40.12 & 30.32 & \underline{54.17} & 47.02 & 85 \\
    GeoMultiTaskNet~\cite{marcocci2023} & 67.53 & 40.86 & 63.89 & 55.31 & 67.02 & \bf{13.85} & \underline{60.97} & 14.08 & 53.09 & 40.33 & 35.02 & \bf{54.79} & 47.22 & 33 \\
    \midrule
    Our Net-C1 & \underline{74.96} & \underline{49.92} & \underline{70.34} & 50.61 & \bf{72.49} & 10.99 & 58.96 & \bf{25.55} & \bf{69.58} & \underline{43.75} & \bf{42.42} & 44.78 & \bf{51.19} & \bf{1.28} \\
    Our Net-E1 & \bf{75.21} & \bf{51.17} & \bf{70.44} & 52.26 & \underline{68.95} & \underline{11.21} & \bf{61.25} & \underline{25.31} & 51.62 & \bf{44.38} & \underline{40.16} & 47.97 & \underline{49.99} & \underline{1.65} \\
    \bottomrule
\end{tabular}}
\vspace{-1mm}
\captionsetup{width=0.96\linewidth}
\caption{Comparison with other UDA methods on the considered target domain test set of the FLAIR \#1 dataset. The discovered lightweight networks improved the performance of the handcrafted baselines, which shows a strong transferability of the discovered networks. Here, all the baselines results were reported in GeoMultiTaskNet~\cite{marcocci2023}. The best score is in \textbf{bold} and the second-best is \underline{underlined}. (Note: \textit{Build=Building}, \textit{Perv.=Pervious}, \textit{Imp.=Impervious}, \textit{Bare=Bare soil}, \textit{Conif.=Coniferous}, \textit{Decid.=Deciduous}, \textit{Brush=Brushwood}, \textit{Vine=Vineyard}, \textit{Herb=Herbaceous}, \textit{Agric=Agriculture land}, and \textit{Plowed=Plowed land})}
\label{tab:flair_transfer_results}
\vspace{0.1cm}
\end{table*}

\begin{figure*}[!t]
\centering
    \includegraphics[width=1\linewidth]{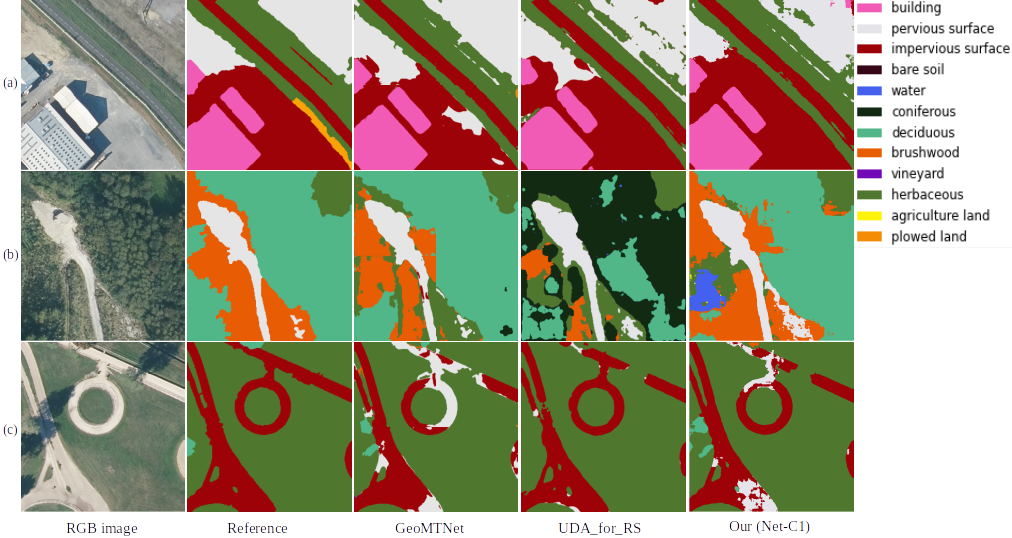}
    \vspace{-5mm}
\captionsetup{width=0.98\linewidth}
\caption{Visual comparison of land cover mapping results of the best-discovered network and some representative baselines in Table~\ref{tab:flair_transfer_results}. The land cover maps of the baselines, GeoMTNet and UDA\_for\_RS, were obtained from GeoMultiTaskNet~\cite{marcocci2023}.}
\label{fig:flair_uda_visual}
\vspace{-0.3cm}
\end{figure*}

\subsection{Searching on OpenEarthMap}\label{sec:5.1}
The Net-C1 and Net-C2 in Table~\ref{tab:oem_search_results} are the top 2 networks discovered by applying confidence-based pseudo-labelling, whereas the Net-E1 and Net-E2 are top 2 discovered by applying energy-based pseudo-labelling. As shown in the Table~\ref{tab:oem_search_results}, with the lowest number of parameters (1.28$M$), Net-C1 outperformed both Net-E1 and Net-E2 by 2.23\% (mIoU). Also, Net-C2 surpassed both Net-E1 and Net-E2 by 1.62\% (mIoU). This shows that the networks discovered via confidence-based pseudo-labelling achieve better performance compared to the ones discovered via energy-based pseudo-labelling. However, we need to mention that the tunable threshold $\tau_e$ and the temperature $T$ of the energy function were not tuned. Tuning them might have improved the performance of the energy-based pseudo-labelling. 

Compared with the baseline methods based on the four metrics we employed, except for the speed (FPS) that IAST and CBST topped, the discovered lightweight networks achieve better performance over all the baselines. This is because the U-Net-based architecture of discovered networks has multiple skip connections, which are good for the performance of the networks, but are detrimental to the speed of the networks. Using fewer parameters (1.28$M$) with only 30.16$G$ FLOPs, the best-discovered network (Net-C1) achieved better accuracy results (59.38\% mIoU) than all the representative baselines. For example, it can be seen from the Tabel~\ref{tab:oem_search_results} that Net-C1 improved the performance (mIoU) of IAST and CBST by 5.92\% and 7.27\%, respectively. Furthermore, Net-C1 demonstrated a big performance gap with regard to AdaptSegNet (13.78\% mIoU) and CLAN (10.53\% mIoU). Focusing on individual classes, it can be observed that Net-C1 has the best performance in most of the classes, except for \textit{bareland}, \textit{road}, and \textit{agriculture}. IAST gained approximately 5.5\% in IoU over Net-C1 for both \textit{bareland} and \textit{agriculture}. The overall results of the discovered lightweight networks show that the proposed method achieves an encouraging performance.

\vspace{-0.5cm}
\paragraph{Visualization:}
In Figure~\ref{fig:oem_uda_visual}, we present examples of land cover mapping results obtained from the best-discovered network (Net-C1) and representative baselines on the OpenEarthMap regional target domain test set (see Table~\ref{tab:oem_search_results}). The Net-C1 produces detailed land cover maps. In row (a), except for some portion of \textit{agricultural land} that Net-C1 misclassified as \textit{rangeland}, IAST and CBST wrongly classified most of the \textit{agricultural land} as \textit{rangeland}. This might be because of the spectra similarity of \textit{agricultural land} and \textit{rangeland}. Although Net-C1 produces the best land cover map in row (b), it failed to identify the \textit{water} at the top-left part of the image, which CBST failed as well but it was identified by IAST. However, in row (c), Net-C1 was able to identify the tiny \textit{water} at the top-right, whereas IAST failed to do so. In all, most of the \textit{buildings} and \textit{roads} were wrongly classified as \textit{developed space} by IAST and CBST because some \textit{developed space} might have cover materials quite similar to \textit{buildings} and \textit{roads}. 

\subsection{Transferrability on FLAIR}\label{sec:5.2}
We also employ the discovered lightweight networks on the challenging FLAIR \#1 dataset. Compared to the 8 classes in the OpenEarthMap dataset, the FLAIR \#1 has 12 classes, which makes it a more difficult benchmark than the OpenEarthMap. Here, we aim to find out if the discovered networks have UDA-oriented capability when they are used on different benchmark datasets. In other words, can the discovered lightweight networks achieve satisfactory performance if they are employed on datasets that they were not searched on? For each of the pseudo-labelling schemes that we employed, only the top network was evaluated on the considered target domain test set of FLAIR \#1. 

As presented in Table~\ref{tab:flair_transfer_results}, using fewer parameters (less than $2M$), both Net-C1 and Net-E1 of the proposed method improved the performance of all the representative baselines results that were reported in GeoMultiTaskNet~\cite{marcocci2023}, the state-of-the-art UDA method on FLAIR \#1. For example, Net-C1 (51.19\% mIoU) and Net-E1 (49.99\% mIoU) improved the performance of GeoMultiTaskNet (47.22\% mIoU) and UDA\_for\_RS (47.02\% mIoU) by approximately 4.0\% and 2.7\%, respectively. The encouraging results of the discovered networks is due to the U-Net-based architecture (which has proven to be good for semantic segmentation) that was employed as the base architecture for the search. For individual class IoU, the Net-C1 performed best in four classes (\textit{water}, \textit{brushwood}, \textit{vineyard} and \textit{agriculture}), and Net-E1 in five classes (\textit{building}, \textit{pervious}, \textit{impervious}, \textit{deciduous}, and \textit{herbaceous}). The GeoMultiTaskNet achieved the best IoUs in \textit{coniferous} and \textit{plowed land}, whereas UDA\_for\_RS in \textit{bare soil}. Overall, the results on the considered target domain test set of FLAIR \#1 show that the discovered lightweight networks have UDA-oriented capability to transfer the knowledge learnt from a source domain to a target domain.

\vspace{-0.3cm}
\paragraph{Visualization:}
Figure~\ref{fig:flair_uda_visual} presents some examples of land cover mapping results on the test set of the FLAIR \#1 target domain that we considered. The maps are obtained from the best-discovered network (Net-C1) and some of the representative baselines provided in Table~\ref{tab:flair_transfer_results}. Net-C1 and UDA\_for\_RS misclassified some portion of \textit{pervious surface} as \textit{herbaceous} in the row (a), however, GeoMTNet wrongly classified some \textit{impervious surface} as \textit{pervious surface}. In row (b), while Net-C1 wrongly classified some \textit{deciduous} as \textit{water} and \textit{brushwood} as \textit{herbaceous}, GeoMTNet misclassified most of the \textit{brushwood} as \textit{herbaceous}. UDA\_for\_RS produced the most misclassification of all the classes. For example, \textit{deciduous} was easily confused with \textit{coniferous} and \textit{brushwood} was confused with \textit{herbaceous}. This might be due to spatial pattern similarities among the images as alluded by GeoMultiTaskNet~\cite{marcocci2023}. In row (c), UDA\_for\_RS produced the best result for the \textit{impervious surface} but failed to identify the \textit{deciduous} at the centre-left of the image. Net-C1 was able to identify the \textit{deciduous}, but it misclassified some \textit{herbaceous} as \textit{deciduous}. GeoMTNet wrongly classified most \textit{impervious surface} as \textit{pervious surface}. The \textit{impervious surface} and \textit{pervious surface} might be confused easily due to their spectra similarities.

\subsection{Limitations}
Although the proposed method demonstrated an encouraging performance it has some limitations. First, the overhead of the Gibbs sampling and inference over a complex loopy network. Hence, the inference has to be approximated as it is normally done in MRF~\cite{koller2009probabilistic} to minimise the cost of inference, which might have affected the optimal solution. Second, the speed of the discovered networks might not be suitable for real-time applications. It would make sense to improve the speed of the discovered networks, maybe by pruning the unnecessary skip connections in the network. Finally, we partially answer the question: "\textit{What pseudo-labelling technique is best suited for UDA-NAS}?". Our observations indicate that confidence-based pseudo-labelling is a good choice. However, we did not tune the hyperparameters of the energy-based pseudo-labelling, hence, the need for further studies in this direction.

\section{Conclusion}\label{sec:6}
The existing UDA methods for land cover mapping are resource-hungry systems, and they are not suitable for real-world applications in resource-constrained platforms. In this paper, we introduce a novel approach to search for lightweight neural networks under a limited resource budget for UDA land cover mapping. In particular, we leverage a self-training UDA framework with a Markov random filed architecture search strategy to learn lightweight architectures that can transfer knowledge learned from a source domain to a target domain. We demonstrate the effectiveness of the proposed approach on two recent UDA benchmark datasets of remote sensing, which improve the performance of several competing UDA methods. We are hoping the UDA-NAS framework proposed here will also suggest something inspiring for the research on UDA-NAS for land cover mapping and remote sensing in general.

{
\small
\bibliographystyle{ieeenat_fullname}
\bibliography{main}

\begin{thebibliography}{78}
\providecommand{\natexlab}[1]{#1}
\providecommand{\url}[1]{\texttt{#1}}
\expandafter\ifx\csname urlstyle\endcsname\relax
  \providecommand{\doi}[1]{doi: #1}\else
  \providecommand{\doi}{doi: \begingroup \urlstyle{rm}\Url}\fi

\bibitem[Adriano et~al.(2021)Adriano, Yokoya, Xia, and Baier]{adriano2021}
Bruno Adriano, Naoto Yokoya, Junshi Xia, and Gerald Baier.
\newblock \emph{{Big Earth Observation Data Processing for Disaster Damage Mapping}}, pages 99--118.
\newblock Springer International Publishing, 2021.

\bibitem[Araslanov and Roth(2021)]{Araslanov2021DASAC}
Nikita Araslanov and Stefan Roth.
\newblock Self-supervised augmentation consistency for adapting semantic segmentation.
\newblock In \emph{Proceedings of the IEEE/CVF Conference on CVPR}, pages 15384--15394, 2021.

\bibitem[Ardywibowo et~al.(2020)Ardywibowo, Boluki, Gong, Wang, and Qian]{ardywibowo2020}
Randy Ardywibowo, Shahin Boluki, Xinyu Gong, Zhangyang Wang, and Xiaoning Qian.
\newblock {NADS: Neural architecture distribution search for uncertainty awareness}.
\newblock In \emph{Proceedings of the 37th International Conference on Machine Learning}, 2020.

\bibitem[B et~al.(2023)B, Sanyal, and Babu]{Prasanna2023CVPR}
Prasanna B, Sunandini Sanyal, and R.~Venkatesh Babu.
\newblock Continual domain adaptation through pruning-aided domain-specific weight modulation.
\newblock In \emph{Proceedings of the IEEE/CVF Conference on CVPR Workshops}, pages 2456--2462, 2023.

\bibitem[Bender et~al.(2018)Bender, Kindermans, Zoph, Vasudevan, and Le]{bender2018understanding}
Gabriel Bender, Pieter-Jan Kindermans, Barret Zoph, Vijay Vasudevan, and Quoc Le.
\newblock Understanding and simplifying one-shot architecture search.
\newblock In \emph{International Conference on Machine Learning}, pages 550--559, 2018.

\bibitem[Berman et~al.(2020)Berman, Pishchulin, Xu, Blaschko, and Medioni]{berman2020aows}
Maxim Berman, Leonid Pishchulin, Ning Xu, Matthew~B Blaschko, and G{\'e}rard Medioni.
\newblock {AOWS}: Adaptive and optimal network width search with latency constraints.
\newblock In \emph{Proceedings of the IEEE/CVF Conference on Computer Vision and Pattern Recognition}, pages 11217--11226, 2020.

\bibitem[Broni-Bediako et~al.(2022)Broni-Bediako, Murata, Mormille, and Atsumi]{bbc2022}
Clifford Broni-Bediako, Yuki Murata, Luiz~H Mormille, and Masayasu Atsumi.
\newblock Evolutionary nas for aerial image segmentation with gene expression programming of cellular encoding.
\newblock \emph{Neural Computing and Applications}, 34:\penalty0 14185--14204, 2022.

\bibitem[Cai et~al.(2022)Cai, Yang, Zheng, Shen, Shang, Yin, and Shi]{caiyu2022}
Yuxiang Cai, Yingchun Yang, Qiyi Zheng, Zhengwei Shen, Yongheng Shang, Jianwei Yin, and Zhongtian Shi.
\newblock {BiFDANet: Unsupervised Bidirectional Domain Adaptation for Semantic Segmentation of Remote Sensing Images}.
\newblock \emph{Remote Sensing}, 14\penalty0 (1), 2022.

\bibitem[Capliez et~al.(2023{\natexlab{a}})Capliez, Ienco, Gaetano, Baghdadi, and Salah]{10089508}
Emmanuel Capliez, Dino Ienco, Raffaele Gaetano, Nicolas Baghdadi, and Adrien~Hadj Salah.
\newblock Temporal-domain adaptation for satellite image time-series land-cover mapping with adversarial learning and spatially aware self-training.
\newblock \emph{IEEE Journal of Selected Topics in Applied Earth Observations and Remote Sensing}, 16:\penalty0 3645--3675, 2023{\natexlab{a}}.

\bibitem[Capliez et~al.(2023{\natexlab{b}})Capliez, Ienco, Gaetano, Baghdadi, Salah, Le~Goff, and Chouteau]{capliez2023}
Emmanuel Capliez, Dino Ienco, Raffaele Gaetano, Nicolas Baghdadi, Adrien~Hadj Salah, Matthieu Le~Goff, and Florient Chouteau.
\newblock Multisensor temporal unsupervised domain adaptation for land cover mapping with spatial pseudo-labeling and adversarial learning.
\newblock \emph{IEEE Transactions on Geoscience and Remote Sensing}, 61:\penalty0 1--16, 2023{\natexlab{b}}.

\bibitem[Chakraborty et~al.(2023)Chakraborty, Sahoo, Panda, and Das]{chakraborty2023anyda}
Omprakash Chakraborty, Aadarsh Sahoo, Rameswar Panda, and Abir Das.
\newblock Any{DA}: Anytime domain adaptation.
\newblock In \emph{The Eleventh International Conference on Learning Representations}, 2023.

\bibitem[Chen et~al.(2023)Chen, Lan, Song, Broni-Bediako, Xia, and Yokoya]{chen2023land}
Hongruixuan Chen, Cuiling Lan, Jian Song, Clifford Broni-Bediako, Junshi Xia, and Naoto Yokoya.
\newblock Land-cover change detection using paired openstreetmap data and optical high-resolution imagery via object-guided transformer, 2023.
\newblock arXiv:2310.02674.

\bibitem[Chen et~al.(2021)Chen, Zhai, Chen, Fang, Zhou, and Yu]{rs13081493}
Jifa Chen, Guojun Zhai, Gang Chen, Bo Fang, Ping Zhou, and Nan Yu.
\newblock Unsupervised domain adaption for high-resolution coastal land cover mapping with category-space constrained adversarial network.
\newblock \emph{Remote Sensing}, 13\penalty0 (8), 2021.

\bibitem[Chu et~al.(2021)Chu, Zhang, and Xu]{chu2021fairnas}
Xiangxiang Chu, Bo Zhang, and Ruijun Xu.
\newblock Fairnas: Rethinking evaluation fairness of weight sharing neural architecture search.
\newblock In \emph{Proceedings of the IEEE/CVF International Conference on Computer Vision}, pages 12239--12248, 2021.

\bibitem[Denby and Lucia(2019)]{denby2019}
Bradley Denby and Brandon Lucia.
\newblock Orbital edge computing: Machine inference in space.
\newblock \emph{IEEE Computer Architecture Letters}, 18\penalty0 (1):\penalty0 59--62, 2019.

\bibitem[Deng et~al.(2021)Deng, Zhu, Tian, and Newsam]{deng2021autoadapt}
Xueqing Deng, Yi Zhu, Yuxin Tian, and Shawn Newsam.
\newblock Autoadapt: Automated segmentation network search for unsupervised domain adaptation, 2021.
\newblock {arXiv:2106.13227}.

\bibitem[Elsken et~al.(2019)Elsken, Metzen, and Hutter]{thomas2019NAS}
Thomas Elsken, Jan~Hendrik Metzen, and Frank Hutter.
\newblock Neural architecture search: a survey.
\newblock \emph{J. Mach. Learn. Res.}, 20\penalty0 (1):\penalty0 1997–2017, 2019.

\bibitem[Garioud et~al.(2022)Garioud, Peillet, Bookjans, Giordano, and Wattrelos]{garioud2022flair1}
Anatol Garioud, Stéphane Peillet, Eva Bookjans, Sébastien Giordano, and Boris Wattrelos.
\newblock {FLAIR \#1: Semantic segmentation and domain adaptation dataset}, 2022.
\newblock arXiv:2211.12979v5.

\bibitem[Guo et~al.(2020)Guo, Zhang, Mu, Heng, Liu, Wei, and Sun]{guo2020single}
Zichao Guo, Xiangyu Zhang, Haoyuan Mu, Wen Heng, Zechun Liu, Yichen Wei, and Jian Sun.
\newblock Single path one-shot neural architecture search with uniform sampling.
\newblock In \emph{{Computer Vision--ECCV 2020}}, pages 544--560, 2020.

\bibitem[Hoyer et~al.(2022)Hoyer, Dai, and Van~Gool]{hoyer2022daformer}
Lukas Hoyer, Dengxin Dai, and Luc Van~Gool.
\newblock {DAFormer}: Improving network architectures and training strategies for domain-adaptive semantic segmentation.
\newblock In \emph{CVPR}, 2022.

\bibitem[Jang et~al.(2017)Jang, Gu, and Poole]{jang2017categorical}
Eric Jang, Shixiang Gu, and Ben Poole.
\newblock Categorical reparameterization with gumbel-softmax.
\newblock In \emph{International Conference on Learning Representations}, 2017.

\bibitem[Ji et~al.(2021)Ji, Wang, and Luo]{9198144}
Shunping Ji, Dingpan Wang, and Muying Luo.
\newblock Generative adversarial network-based full-space domain adaptation for land cover classification from multiple-source remote sensing images.
\newblock \emph{IEEE Transactions on Geoscience and Remote Sensing}, 59\penalty0 (5):\penalty0 3816--3828, 2021.

\bibitem[Jia et~al.(2024)Jia, Chen, Zhang, Sang, and Zhang]{JIA202497}
Peiyan Jia, Chen Chen, Delong Zhang, Yulong Sang, and Lei Zhang.
\newblock {Semantic segmentation of deep learning remote sensing images based on band combination principle: Application in urban planning and land use}.
\newblock \emph{Computer Communications}, 217:\penalty0 97--106, 2024.

\bibitem[Kapil et~al.(2024)Kapil, Marvasti-Zadeh, Erbilgin, and Ray]{kapil2024shadowsense}
Rudraksh Kapil, Seyed~Mojtaba Marvasti-Zadeh, Nadir Erbilgin, and Nilanjan Ray.
\newblock Shadowsense: Unsupervised domain adaptation and feature fusion for shadow-agnostic tree crown detection from rgb-thermal drone imagery.
\newblock In \emph{Proceedings of the IEEE/CVF Winter Conference on Applications of Computer Vision}, pages 8266--8276, 2024.

\bibitem[Koller and Friedman(2009)]{koller2009probabilistic}
D. Koller and N. Friedman.
\newblock \emph{Probabilistic Graphical Models: Principles and Techniques}.
\newblock MIT Press, 2009.

\bibitem[LeCun et~al.(2006)LeCun, Chopra, Hadsell, Ranzato, and Huang]{lecun2006tutorial}
Yann LeCun, Sumit Chopra, Raia Hadsell, M Ranzato, and Fujie Huang.
\newblock A tutorial on energy-based learning.
\newblock \emph{Predicting structured data}, 1\penalty0 (0), 2006.

\bibitem[Li et~al.(2022)Li, Gao, Su, and Momanyi]{li2022}
Weitao Li, Hui Gao, Yi Su, and Biffon~Manyura Momanyi.
\newblock {Unsupervised Domain Adaptation for Remote Sensing Semantic Segmentation with Transformer}.
\newblock \emph{Remote Sensing}, 14\penalty0 (19), 2022.

\bibitem[Li and Peng(2020)]{li2020network}
Yichen Li and Xingchao Peng.
\newblock Network architecture search for domain adaptation, 2020.
\newblock {arXiv:2008.05706}.

\bibitem[Li et~al.(2020)Li, Yang, Wang, and Xu]{Yanxi2020}
Yanxi Li, Zhaohui Yang, Yunhe Wang, and Chang Xu.
\newblock {Adapting Neural Architectures Between Domains}.
\newblock In \emph{Advances in Neural Information Processing Systems}, pages 789--798, 2020.

\bibitem[Liang et~al.(2023)Liang, Cheng, Xiao, and Dong]{liang2023}
Chenbin Liang, Bo Cheng, Baihua Xiao, and Yunyun Dong.
\newblock Unsupervised domain adaptation for remote sensing image segmentation based on adversarial learning and self-training.
\newblock \emph{IEEE Geoscience and Remote Sensing Letters}, 20:\penalty0 1--5, 2023.

\bibitem[Liu et~al.(2019)Liu, Simonyan, and Yang]{liu2018darts}
Hanxiao Liu, Karen Simonyan, and Yiming Yang.
\newblock {DARTS}: Differentiable architecture search.
\newblock In \emph{International Conference on Learning Representations}, 2019.

\bibitem[Liu et~al.(2022{\natexlab{a}})Liu, Yoo, Xing, Oh, El~Fakhri, Kang, Woo, et~al.]{liu2022deepUDA}
Xiaofeng Liu, Chaehwa Yoo, Fangxu Xing, Hyejin Oh, Georges El~Fakhri, Je-Won Kang, Jonghye Woo, et~al.
\newblock Deep unsupervised domain adaptation: A review of recent advances and perspectives.
\newblock \emph{APSIPA Transactions on Signal and Information Processing}, 11\penalty0 (1), 2022{\natexlab{a}}.

\bibitem[Liu et~al.(2022{\natexlab{b}})Liu, Kang, Huang, Wang, and Yang]{yikun2022}
Yikun Liu, Xudong Kang, Yuwen Huang, Kuikui Wang, and Gongping Yang.
\newblock Unsupervised domain adaptation semantic segmentation for remote-sensing images via covariance attention.
\newblock \emph{IEEE Geoscience and Remote Sensing Letters}, 19:\penalty0 1--5, 2022{\natexlab{b}}.

\bibitem[Liu et~al.(2023)Liu, Sun, Xue, Zhang, Yen, and Tan]{liuNAS2023}
Yuqiao Liu, Yanan Sun, Bing Xue, Mengjie Zhang, Gary~G. Yen, and Kay~Chen Tan.
\newblock A survey on evolutionary neural architecture search.
\newblock \emph{IEEE Transactions on Neural Networks and Learning Systems}, 34\penalty0 (2):\penalty0 550--570, 2023.

\bibitem[Luo and Ji(2022)]{LUO2022105}
Muying Luo and Shunping Ji.
\newblock Cross-spatiotemporal land-cover classification from vhr remote sensing images with deep learning based domain adaptation.
\newblock \emph{ISPRS Journal of Photogrammetry and Remote Sensing}, 191:\penalty0 105--128, 2022.

\bibitem[Luo et~al.(2019)Luo, Zheng, Guan, Yu, and Yang]{luo2019Taking}
Yawei Luo, Liang Zheng, Tao Guan, Junqing Yu, and Yi Yang.
\newblock Taking a closer look at domain shift: Category-level adversaries for semantics consistent domain adaptation.
\newblock In \emph{CVPR}, 2019.

\bibitem[Ma et~al.(2023)Ma, Zheng, Wang, and Zhong]{10094018}
Ailong Ma, Chenyu Zheng, Junjue Wang, and Yanfei Zhong.
\newblock Domain adaptive land-cover classification via local consistency and global diversity.
\newblock \emph{IEEE Transactions on Geoscience and Remote Sensing}, 61:\penalty0 1--17, 2023.

\bibitem[Marsocci et~al.(2023)Marsocci, Gonthier, Garioud, Scardapane, and Mallet]{marcocci2023}
V. Marsocci, N. Gonthier, A. Garioud, S. Scardapane, and C. Mallet.
\newblock Geomultitasknet: remote sensing unsupervised domain adaptation using geographical coordinates.
\newblock In \emph{2023 IEEE/CVF Conference on Computer Vision and Pattern Recognition Workshops (CVPRW)}, pages 2075--2085, 2023.

\bibitem[Mei et~al.(2020)Mei, Zhu, Zou, and Zhang]{mei2020instance}
Ke Mei, Chuang Zhu, Jiaqi Zou, and Shanghang Zhang.
\newblock Instance adaptive self-training for unsupervised domain adaptation.
\newblock In \emph{ECCV}, 2020.

\bibitem[Melis et~al.(2018)Melis, Dess\`\i, Loddo, {La Mantia}, {Da Pelo}, Deflorio, Ghiglieri, Hailu, Kalegele, and Mwasi]{melis2018}
M~T Melis, F Dess\`\i, P Loddo, C {La Mantia}, S {Da Pelo}, A~M Deflorio, G Ghiglieri, B~T Hailu, K Kalegele, and B~N Mwasi.
\newblock Improving land cover mapping: A mobile application based on esa sentinel-2 imagery.
\newblock \emph{The International Archives of the Photogrammetry, Remote Sensing and Spatial Information Sciences}, XLII-3:\penalty0 1263--1266, 2018.

\bibitem[Meng et~al.(2022)Meng, Chen, Yang, Song, Lin, Xie, Pu, Wang, Song, and Zhuang]{meng2022slimmable}
Rang Meng, Weijie Chen, Shicai Yang, Jie Song, Luojun Lin, Di Xie, Shiliang Pu, Xinchao Wang, Mingli Song, and Yueting Zhuang.
\newblock Slimmable domain adaptation.
\newblock In \emph{Proceedings of the IEEE/CVF Conference on Computer Vision and Pattern Recognition}, pages 7141--7150, 2022.

\bibitem[Mollick et~al.(2023)Mollick, Azam, and Karim]{MOLLICK2023}
Taposh Mollick, Md~Golam Azam, and Sabrina Karim.
\newblock {Geospatial-based machine learning techniques for land use and land cover mapping using a high-resolution unmanned aerial vehicle image}.
\newblock \emph{Remote Sensing Applications: Society and Environment}, 29:\penalty0 100859, 2023.

\bibitem[Niu et~al.(2021)Niu, Wu, Xu, Zhang, Guo, Zhao, Wang, and Tan]{niu2021adaxpert}
Shuaicheng Niu, Jiaxiang Wu, Guanghui Xu, Yifan Zhang, Yong Guo, Peilin Zhao, Peng Wang, and Mingkui Tan.
\newblock Adaxpert: Adapting neural architecture for growing data.
\newblock In \emph{The International Conference on Machine Learning}, 2021.

\bibitem[Olsson et~al.(2021)Olsson, Tranheden, Pinto, and Svensson]{olsson2021classmix}
Viktor Olsson, Wilhelm Tranheden, Juliano Pinto, and Lennart Svensson.
\newblock Classmix: Segmentation-based data augmentation for semi-supervised learning.
\newblock In \emph{Proceedings of the IEEE/CVF Winter Conference on Applications of Computer Vision}, pages 1369--1378, 2021.

\bibitem[Payman et~al.(2012)Payman, Abner, Dhruv, and Yadollahpour]{payman2012}
Payman, Guzman-Rivera Abner, Shakhnarovich Gregory~Batra Dhruv, and Yadollahpour.
\newblock Diverse m-best solutions in markov random fields.
\newblock pages 1--16, 2012.

\bibitem[Qin and Liu(2022)]{qui2022}
Rongjun Qin and Tao Liu.
\newblock {A Review of Landcover Classification with Very-High Resolution Remotely Sensed Optical Images: Analysis Unit, Model Scalability and Transferability}.
\newblock \emph{Remote Sensing}, 14\penalty0 (3), 2022.

\bibitem[Richards et~al.(2022)Richards, Richards, et~al.]{richards2022remote}
John~A Richards, John~A Richards, et~al.
\newblock \emph{Remote sensing digital image analysis}.
\newblock Springer, 2022.

\bibitem[Robbiano et~al.(2022)Robbiano, Rahman, Galasso, Caputo, and Carlucci]{Robbiano2022}
L. Robbiano, M.~Ur Rahman, F. Galasso, B. Caputo, and F. Carlucci.
\newblock Adversarial branch architecture search for unsupervised domain adaptation.
\newblock In \emph{2022 IEEE/CVF Winter Conference on Applications of Computer Vision (WACV)}, pages 1008--1018, 2022.

\bibitem[Ronneberger et~al.(2015)Ronneberger, Fischer, and Brox]{ronneberger2015u}
Olaf Ronneberger, Philipp Fischer, and Thomas Brox.
\newblock {U-Net: Convolutional networks for biomedical image segmentation}.
\newblock In \emph{Medical Image Computing and Computer-Assisted Intervention--MICCAI 2015: 18th International Conference, Munich, Germany, October 5-9, 2015, Proceedings, Part III 18}, pages 234--241, 2015.

\bibitem[Seelan et~al.(2003)Seelan, Laguette, Casady, and Seielstad]{SEELAN2003}
Santhosh~K Seelan, Soizik Laguette, Grant~M Casady, and George~A Seielstad.
\newblock {Remote sensing applications for precision agriculture: A learning community approach}.
\newblock \emph{Remote Sensing of Environment}, 88\penalty0 (1):\penalty0 157--169, 2003.

\bibitem[Sheng et~al.(2021)Sheng, Li, Zheng, Liang, Dong, Huang, Ji, and Sun]{sheng2021evolving}
Kekai Sheng, Ke Li, Xiawu Zheng, Jian Liang, Weiming Dong, Feiyue Huang, Rongrong Ji, and Xing Sun.
\newblock On evolving attention towards domain adaptation, 2021.
\newblock {arXiv:2103.13561}.

\bibitem[{Soto Vega} et~al.(2021){Soto Vega}, da~Costa, Feitosa, {Ortega Adarme}, de~Almeida, Heipke, and Rottensteiner]{SOTOVEGA2021113}
Pedro~Juan {Soto Vega}, Gilson Alexandre Ostwald~Pedro da Costa, Raul~Queiroz Feitosa, Mabel~Ximena {Ortega Adarme}, Claudio~Aparecido de Almeida, Christian Heipke, and Franz Rottensteiner.
\newblock {An unsupervised domain adaptation approach for change detection and its application to deforestation mapping in tropical biomes}.
\newblock \emph{ISPRS Journal of Photogrammetry and Remote Sensing}, 181:\penalty0 113--128, 2021.

\bibitem[Tarvainen and Valpola(2017)]{tarvainen2017mean}
Antti Tarvainen and Harri Valpola.
\newblock Mean teachers are better role models: Weight-averaged consistency targets improve semi-supervised deep learning results.
\newblock \emph{Advances in Neural Information Processing Systems}, 30, 2017.

\bibitem[Tasar et~al.(2020)Tasar, Happy, Tarabalka, and Alliez]{9047180}
Onur Tasar, S.~L. Happy, Yuliya Tarabalka, and Pierre Alliez.
\newblock Colormapgan: Unsupervised domain adaptation for semantic segmentation using color mapping generative adversarial networks.
\newblock \emph{IEEE Transactions on Geoscience and Remote Sensing}, 58\penalty0 (10):\penalty0 7178--7193, 2020.

\bibitem[Tian et~al.(2022)Tian, Mithun, Seymour, Chiu, and Kira]{tian2022}
Junjiao Tian, Niluthpol~Chowdhury Mithun, Zachary Seymour, Han-Pang Chiu, and Zsolt Kira.
\newblock Striking the right balance: Recall loss for semantic segmentation.
\newblock In \emph{{2022 International Conference on Robotics and Automation (ICRA)}}, page 5063–5069, 2022.

\bibitem[Tranheden et~al.(2021)Tranheden, Olsson, Pinto, and Svensson]{tranheden2021dacs}
Wilhelm Tranheden, Viktor Olsson, Juliano Pinto, and Lennart Svensson.
\newblock {DACS: Domain adaptation via cross-domain mixed sampling}.
\newblock In \emph{Proceedings of the IEEE/CVF Winter Conference on Applications of Computer Vision}, pages 1379--1389, 2021.

\bibitem[Tsai et~al.(2018)Tsai, Hung, Schulter, Sohn, Yang, and Chandraker]{Tsai_adaptseg_2018}
Y.-H. Tsai, W.-C. Hung, S. Schulter, K. Sohn, M.-H. Yang, and M. Chandraker.
\newblock Learning to adapt structured output space for semantic segmentation.
\newblock In \emph{IEEE Conference on Computer Vision and Pattern Recognition}, 2018.

\bibitem[Vu et~al.(2019)Vu, Jain, Bucher, Cord, and P{\'e}rez]{vu2018advent}
Tuan-Hung Vu, Himalaya Jain, Maxime Bucher, Mathieu Cord, and Patrick P{\'e}rez.
\newblock Advent: Adversarial entropy minimization for domain adaptation in semantic segmentation.
\newblock In \emph{CVPR}, 2019.

\bibitem[Wang et~al.(2020)Wang, Shen, Zhang, Duan, and Mei]{Haoran_2020_ECCV}
Haoran Wang, Tong Shen, Wei Zhang, Lingyu Duan, and Tao Mei.
\newblock Classes matter: A fine-grained adversarial approach to cross-domain semantic segmentation.
\newblock In \emph{The European Conference on Computer Vision (ECCV)}, 2020.

\bibitem[Wang et~al.(2022)Wang, Ma, Zhong, Zheng, and Zhang]{WANG2022113058}
Junjue Wang, Ailong Ma, Yanfei Zhong, Zhuo Zheng, and Liangpei Zhang.
\newblock Cross-sensor domain adaptation for high spatial resolution urban land-cover mapping: From airborne to spaceborne imagery.
\newblock \emph{Remote Sensing of Environment}, 277:\penalty0 113058, 2022.

\bibitem[Wang et~al.(2019)Wang, Jin, Long, Wang, and Jordan]{Wang19TransNorm}
Ximei Wang, Ying Jin, Mingsheng Long, Jianmin Wang, and Michael~I Jordan.
\newblock Transferable normalization: Towards improving transferability of deep neural networks.
\newblock In \emph{NeurIPS}, 2019.

\bibitem[Wang and Blaschko(2022)]{Wang2022MRF-UNets}
Zifu Wang and Matthew~B. Blaschko.
\newblock {MRF-UNets: Searching UNet with Markov Random Fields}.
\newblock In \emph{{ECML-PKDD}}, 2022.

\bibitem[Wu et~al.(2019)Wu, Dai, Zhang, Wang, Sun, Wu, Tian, Vajda, Jia, and Keutzer]{wu2019fbnet}
Bichen Wu, Xiaoliang Dai, Peizhao Zhang, Yanghan Wang, Fei Sun, Yiming Wu, Yuandong Tian, Peter Vajda, Yangqing Jia, and Kurt Keutzer.
\newblock {FbNet}: Hardware-aware efficient convnet design via differentiable neural architecture search.
\newblock In \emph{Proceedings of the IEEE/CVF Conference on CVPR}, pages 10734--10742, 2019.

\bibitem[Xia et~al.(2023{\natexlab{a}})Xia, Wu, Yao, Zhu, Gong, Yang, Hu, and Mo]{xia2023damage}
Haobin Xia, Jianjun Wu, Jiaqi Yao, Hong Zhu, Adu Gong, Jianhua Yang, Liuru Hu, and Fan Mo.
\newblock {A Deep Learning Application for Building Damage Assessment Using Ultra-High-Resolution Remote Sensing Imagery in Turkey Earthquake}.
\newblock \emph{International Journal of Disaster Risk Science}, 14\penalty0 (6):\penalty0 947--962, 2023{\natexlab{a}}.

\bibitem[Xia et~al.(2023{\natexlab{b}})Xia, Yokoya, Adriano, and Broni-Bediako]{junshi2023}
J. Xia, N. Yokoya, B. Adriano, and C. Broni-Bediako.
\newblock Openearthmap: A benchmark dataset for global high-resolution land cover mapp.
\newblock In \emph{2023 IEEE/CVF Winter Conference on Applications of Computer Vision (WACV)}, pages 6243--6253, 2023{\natexlab{b}}.

\bibitem[Xie et~al.(2018)Xie, Zheng, Liu, and Lin]{xie2018snas}
Sirui Xie, Hehui Zheng, Chunxiao Liu, and Liang Lin.
\newblock {SNAS}: stochastic neural architecture search.
\newblock \emph{arXiv preprint arXiv:1812.09926}, 2018.

\bibitem[Xu et~al.(2022)Xu, Wu, Chen, Zhang, and Guo]{mengqui2022UDAsurvey}
Mengqiu Xu, Ming Wu, Kaixin Chen, Chuang Zhang, and Jun Guo.
\newblock {The Eyes of the Gods: A Survey of Unsupervised Domain Adaptation Methods Based on Remote Sensing Data}.
\newblock \emph{Remote Sensing}, 14\penalty0 (17), 2022.

\bibitem[Yu and Huang(2019)]{yu2019universally}
Jiahui Yu and Thomas~S Huang.
\newblock Universally slimmable networks and improved training techniques.
\newblock In \emph{Proceedings of the IEEE/CVF International Conference on Computer Vision}, pages 1803--1811, 2019.

\bibitem[Yu et~al.(2023)Yu, Li, and Lee]{yu2023inpl}
Zhuoran Yu, Yin Li, and Yong~Jae Lee.
\newblock In{PL}: Pseudo-labeling the inliers first for imbalanced semi-supervised learning.
\newblock In \emph{The Eleventh International Conference on Learning Representations}, 2023.

\bibitem[Yuan et~al.(2020)Yuan, Shen, Li, Li, Li, Jiang, Xu, Tan, Yang, Wang, Gao, and Zhang]{YUAN2020}
Qiangqiang Yuan, Huanfeng Shen, Tongwen Li, Zhiwei Li, Shuwen Li, Yun Jiang, Hongzhang Xu, Weiwei Tan, Qianqian Yang, Jiwen Wang, Jianhao Gao, and Liangpei Zhang.
\newblock {Deep learning in environmental remote sensing: Achievements and challenges}.
\newblock \emph{Remote Sensing of Environment}, 241:\penalty0 111716, 2020.

\bibitem[Yue et~al.(2021)Yue, Guo, and Zhang]{yue2021domain}
Zhixiong Yue, Pengxin Guo, and Yu Zhang.
\newblock Domain adaptation by maximizing population correlation with neural architecture search, 2021.
\newblock {arXiv:2109.06652}.

\bibitem[Yue et~al.(2022)Yue, Guo, Zhang, and Liang]{Yue2022}
Zhixiong Yue, Pengxin Guo, Yu Zhang, and Christy Liang.
\newblock Learning feature alignment architecture for domain adaptation.
\newblock In \emph{2022 International Joint Conference on Neural Networks (IJCNN)}, pages 1--8, 2022.

\bibitem[Zhang et~al.(2021)Zhang, Zhang, Zhang, Chen, Wang, and Wen]{zhang2021prototypical}
Pan Zhang, Bo Zhang, Ting Zhang, Dong Chen, Yong Wang, and Fang Wen.
\newblock Prototypical pseudo label denoising and target structure learning for domain adaptive semantic segmentation.
\newblock In \emph{Proceedings of the IEEE/CVF Conference on CVPR}, pages 12414--12424, 2021.

\bibitem[Zhou et~al.(2022)Zhou, Feng, Gu, Pang, Cheng, Lu, Shi, and Ma]{zhou2022context}
Qianyu Zhou, Zhengyang Feng, Qiqi Gu, Jiangmiao Pang, Guangliang Cheng, Xuequan Lu, Jianping Shi, and Lizhuang Ma.
\newblock Context-aware mixup for domain adaptive semantic segmentation.
\newblock \emph{IEEE Transactions on Circuits and Systems for Video Technology}, 33\penalty0 (2):\penalty0 804--817, 2022.

\bibitem[Zoph and Le(2017)]{zoph2017neural}
Barret Zoph and Quoc Le.
\newblock Neural architecture search with reinforcement learning.
\newblock In \emph{International Conference on Learning Representations}, 2017.

\bibitem[Zoph et~al.(2018)Zoph, Vasudevan, Shlens, and Le]{zoph2018learning}
Barret Zoph, Vijay Vasudevan, Jonathon Shlens, and Quoc~V Le.
\newblock Learning transferable architectures for scalable image recognition.
\newblock In \emph{Proceedings of the IEEE Conference on CVPR}, pages 8697--8710, 2018.

\bibitem[Zou et~al.(2018)Zou, Yu, Kumar, and Wang]{zou2018unsupervised}
Yang Zou, Zhiding Yu, BVK Kumar, and Jinsong Wang.
\newblock Unsupervised domain adaptation for semantic segmentation via class-balanced self-training.
\newblock In \emph{{Proceedings of the European Conference on Computer Vision (ECCV)}}, pages 289--305, 2018.

\bibitem[Zou et~al.(2019)Zou, Yu, Liu, Kumar, and Wang]{zou2019confidence}
Yang Zou, Zhiding Yu, Xiaofeng Liu, BVK Kumar, and Jinsong Wang.
\newblock Confidence regularized self-training.
\newblock In \emph{Proceedings of the IEEE/CVF International Conference on Computer Vision}, pages 5982--5991, 2019.

\end{thebibliography}
}

\balance

\end{document}